\pdfoutput=1
\documentclass[11pt]{article}

\usepackage[]{EACL2023}

\usepackage{times}
\usepackage{latexsym}

\usepackage[T1]{fontenc}
\usepackage[utf8]{inputenc}

\usepackage{microtype}

\usepackage{inconsolata}

\usepackage{booktabs}
\usepackage{amsmath}
\usepackage{amsfonts}
\usepackage{graphicx}
\graphicspath{ {images/} }
\usepackage{color,soul}

\usepackage{cleveref}
\usepackage{commath}
\usepackage{multirow}
\usepackage{hyperref}
\usepackage{textcomp}
\usepackage{siunitx}

\crefformat{section}{\S#2#1#3} % see manual of cleveref, section 8.2.1
\crefformat{subsection}{\S#2#1#3}
\crefformat{subsubsection}{\S#2#1#3}
\usepackage{float}
\DeclareMathOperator*{\argmax}{arg\,max}
\newcommand\tab[1][0.5cm]{\hspace*{#1}}
\usepackage{tikz}
\usepackage{tikz}
\newcommand\checkmarknew[1][]{%
  \tikz[scale=0.4,#1]{\fill(0,.35) -- (.25,0) -- (1,.7) -- (.25,.15) -- cycle;}%
}

\newcommand\crossmark[1][]{%
  \tikz[scale=0.4,#1]{
    \fill(0,0)--(0.1,0) .. controls (0.5,0.4) .. (1,0.7)--(0.9,0.7) ..  controls (0.5,0.5) ..(0,0.1) --cycle;
    \fill(1,0.1)--(0.9,0.1) .. controls (0.5,0.3) .. (0,0.7)--(0.1,0.7) .. controls (0.5,0.4) ..(1,0.2) --cycle;
  }%
}
\usepackage[normalem]{ulem}
\usepackage{xspace} 
\usepackage{enumitem}
\usepackage{pifont}% http://ctan.org/pkg/pifont
\usepackage{comment}
\usepackage{soul}
\usepackage{color}

\definecolor{lightsalmon}{RGB}{255, 230, 230}
\definecolor{darksalmon}{RGB}{255, 194, 194}
\definecolor{mygreen}{RGB}{165, 243, 165}
\definecolor{palegreen}{RGB}{217, 246, 217}
\definecolor{palestgreen}{RGB}{235, 250, 235}
\definecolor{lightblue}{RGB}{204, 243, 255}
\definecolor{theblue}{RGB}{201, 218, 248}
\definecolor{thepurple}{RGB}{218, 202, 255}
\definecolor{vermilion}{RGB}{255, 60, 0}

\usepackage{tcolorbox}
\tcbuselibrary{skins,breakable,xparse}

\usepackage{color, colortbl}
\definecolor{Gray}{gray}{0.9}
\definecolor{LightCyan}{rgb}{0.88,1,1}
\usepackage[first=0,last=9]{lcg}

\newcommand\smalltab[1][2mm]{\hspace*{#1}}
\newcommand{\method}{\textsc{S}ty\textsc{LE}x\xspace}
\newcommand{\features}{features\xspace}

\title{\method: Explaining Style Using Human Lexical Annotations}
\author{Shirley Anugrah Hayati\raisebox{3pt}{{\includegraphics[height=1em,width=1em]{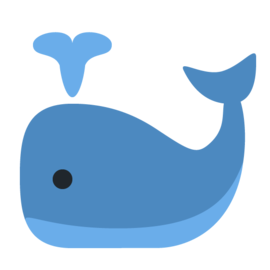}}} \tab Kyumin Park\raisebox{3pt}{{\includegraphics[height=1em,width=1em]{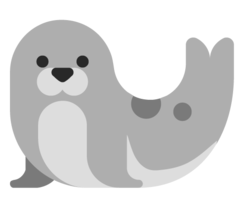}}} \tab Dheeraj Rajagopal\thanks{\smalltab currently at Google}\smalltab 
 \raisebox{3pt}{{\includegraphics[height=1em,width=1em]{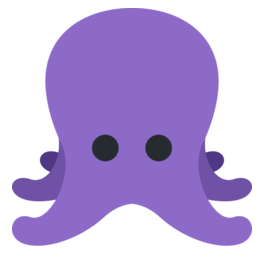}}} \\
\textbf{Lyle Ungar}\raisebox{3pt}{{\includegraphics[height=1em,width=1em]{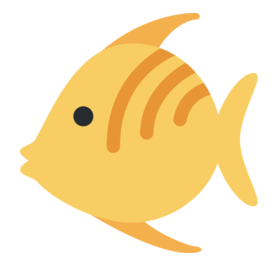}}} \tab \textbf{Dongyeop Kang}\raisebox{3pt}{{\includegraphics[height=1em,width=1em]{emoji/whale.png}}}\\
  \raisebox{3pt}{{\includegraphics[height=1em,width=1em]{emoji/whale.png}}}University of Minnesota \tab  \raisebox{3pt}{{\includegraphics[height=1em,width=1em]{emoji/seal.png}}}KAIST \\
  \tab \raisebox{3pt}{{\includegraphics[height=1em,width=1em]{emoji/octopus.png}}}Carnegie Mellon University \tab \raisebox{3pt}{{\includegraphics[height=1em,width=1em]{emoji/yellowfish.png}}}University of Pennsylvania\\
{\sffamily hayat023@umn.edu} \tab
\textsf{pkm9403@kaist.ac.kr} \tab
\textsf{dheeraj@cs.cmu.edu} \tab \\ \textsf{ungar@cis.upenn.edu} \tab \textsf{dongyeop@umn.edu}\\}

\begin{document}
\maketitle
\begin{abstract}
Large pre-trained language models have achieved impressive results on various style classification tasks, but they often learn spurious domain-specific words to make predictions \cite{hayati-etal-2021-bert}. 
While human explanation highlights stylistic tokens as important features for this task, we observe that model explanations often do not align with them. 
To tackle this issue, we introduce \textbf{\method}, a model that learns from human-annotated explanations of stylistic \features and jointly learns to perform the task and predict these \features as model explanations. 
Our experiments show that \method can provide human-like stylistic lexical explanations without sacrificing the performance of sentence-level style prediction on both in-domain and out-of-domain datasets. Explanations from \method show significant improvements in explanation metrics (sufficiency, plausibility) and when evaluated with human annotations. They are also more understandable by human judges compared to the widely-used saliency-based explanation baseline.\footnote{Code and data are publicly available at \url{https://github.com/minnesotanlp/stylex}}
\end{abstract}

\section{Introduction}
\begin{figure}
    \centering
    \includegraphics[trim=0cm 9cm 0cm 0cm,clip,width=\linewidth]{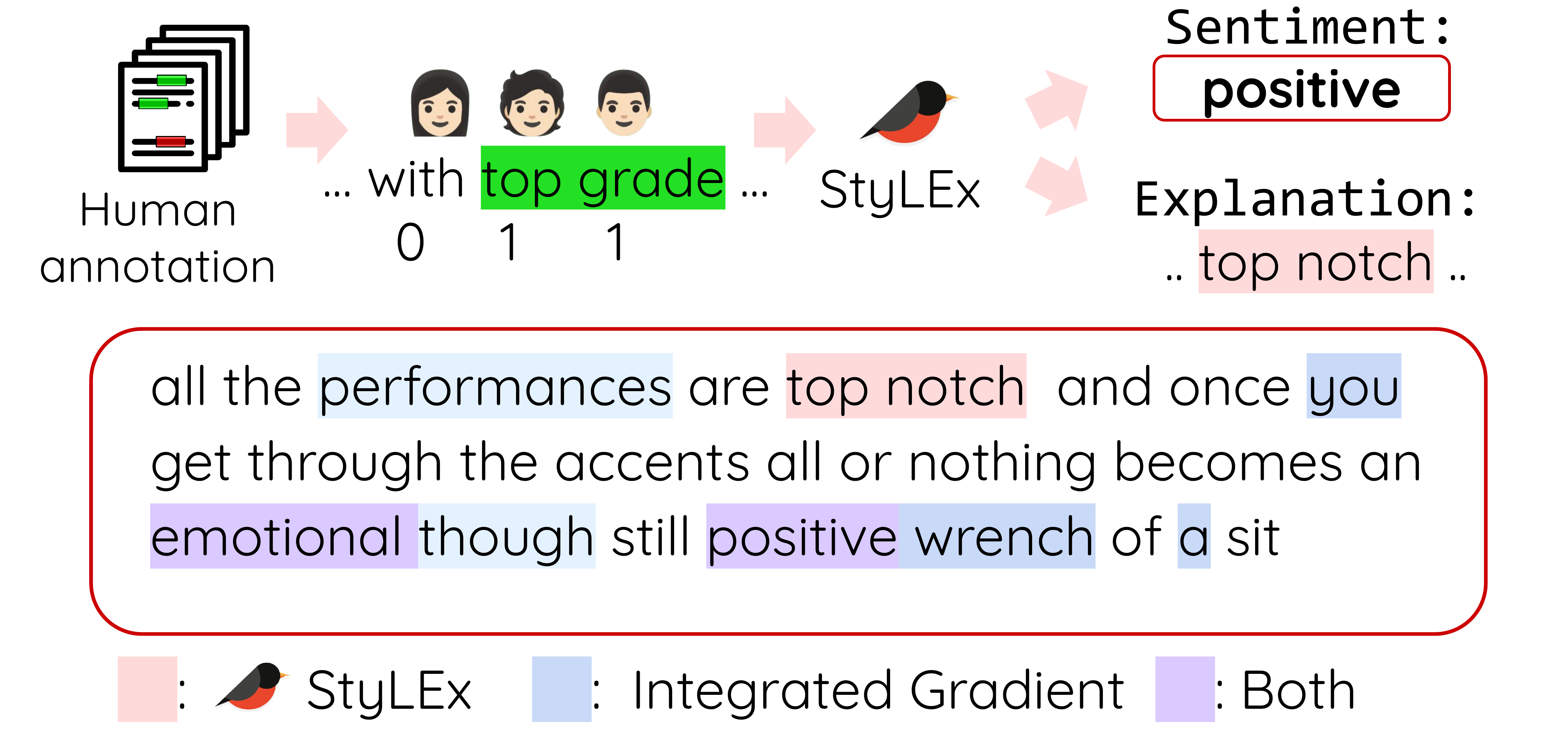}
    \caption{
    % StyLEx uses \textcolor{red}{human perception scores} for classifying the input sentence's style and provides lexical explanation. 
    StyLEx classifies the input sentence's style and provides lexical explanation. 
    % We train StyLEx using human-annotated stylistic cues from \citet{hayati-etal-2021-bert}.
    Compared to explanations computed by the integrated gradient method \cite{mudrakarta-etal-2018-model}, \method can find more accurate stylistic words.
    % and provide broader coverage of contextual stylistic lexica. 
    \colorbox{green!40}{Green} highlight refers to human's annotated positive word, \colorbox{darksalmon!40}{pink} for \method, \colorbox{theblue!50}{blue} for baseline, and \colorbox{thepurple!40}{purple} for both \method and the baseline. 
    \vspace{-4mm}
    }
    \label{fig:task example}
\end{figure}

People use style as a strategic choice for their personal or social goals in communications, making style analysis a long-studied field in NLP \cite{ hovy1987generating, kabbara2016stylistic,kang-hovy-2021-style}. While large language models have obtained state-of-the-art results on many NLP tasks, they have been shown to overfit to spurious correlations in data across several datasets \cite{sen-etal-2021-counterfactually,schlangen-2021-targeting, bras-v119-bras20a}. \citet{hayati-etal-2021-bert} found a phenomenon in style classification tasks where the model's word-level explanation do not align with human's stylistic cues (stylistic cues are words that signify the style of a text). 
For instance, words such as ``performances'' and ``wrench'' in Figure \ref{fig:task example} are marked as important cues for sentiment by a saliency method. However, they are different from words that humans perceive as essential \features for predicting the style (``top,'' ``notch''). 
% Thus, it motivates us to align the model's attention toward the important words that humans use in identifying the style of a text. 
% In this study, we introduce {\bf \method}, a style classification model that jointly learns to align human-annotated stylistic cues as explanations and then predict the style of the overall sentence based on the cues (Figure \ref{fig:task example}). 

% In the past, researchers have developed stylistic lexicon dictionary and identified the style of a text, such as sentiment or emotion,  based on the presence of words in the dictionary \cite{mohammad-turney-2010-emotions, luis2021, hutto2014vader, liwc_tausczik2010psychological}. Traditional dictionary-based lexicon matching methods \cite{taboada2011lexicon, eisenstein2017unsupervised} could provide high precision and interpretability for detecting these stylistic cues, but they lack coverage and context awareness.
% On the other hand, contextualized language models, such as BERT \cite{devlin-etal-2019-bert}, are supposed to detect important words within a given context dynamically. However, they are treated as an opaque model and sometimes are biased toward the training data domain \cite{hayati-etal-2021-bert}.
% Humans often rely on identifying the stylistic words from a text and then predict the sentence style (e.g., hedges for politeness, swear words for offensiveness) \cite{danescu-niculescu-mizil-etal-2013-computational, luis2021}. 
Prior research in style have developed stylistic lexicon dictionary to identify the style of a text, such as sentiment or emotion, and in-turn incorporated them for style classification tasks \cite{mohammad-turney-2010-emotions, hutto2014vader, liwc_tausczik2010psychological}. While lexicon-based matching methods \cite{taboada2011lexicon, eisenstein2017unsupervised} provide interpretability for the task, they lack coverage and do not incorporate the context for prediction. On the other hand, current large scale models like BERT \cite{devlin-etal-2019-bert} are effective at style classification. However, their explanations often reveal that the model do not rely on the stylistic words to make the prediction. In this work, we hypothesize that leveraging stylistic lexica along with the effectiveness of a large scale model like BERT can not only predict style but also provide meaningful explanations that align with human style explanations. 
% Past research have developed stylistic lexicon dictionary and identified the style of a text, such as sentiment or emotion,  based on the presence of words in the dictionary \cite{mohammad-turney-2010-emotions, luis2021, hutto2014vader, liwc_tausczik2010psychological}. Traditional dictionary-based lexicon matching methods \cite{taboada2011lexicon, eisenstein2017unsupervised} could provide high precision and interpretability for detecting these stylistic cues, but they lack coverage and context awareness. On the other hand, recent contextualized language models, such as BERT \cite{devlin-etal-2019-bert}, perform really well on sentence-level tasks but their important words do not align with those from humans.
% are supposed to detect important words within a given context dynamically. However, they are treated as an opaque model and sometimes are biased toward the training data domain \cite{hayati-etal-2021-bert}.

Towards this, we introduce {\bf \method}, a style classification model that jointly learns to align human-annotated stylistic cues as explanations and then predict the style of the overall sentence based on the cues (Figure \ref{fig:task example}). % Based on our intuition that in identifying the style of a text, humans often look at the important words that signify the style, such as hedges for politeness or swear words for offensiveness. We developed \method to mimic this behavior, and automatically expand the lexicon set (Figure \ref{fig:task example} bottom). Obtaining human lexical annotation in a large-scale training data, however, is very costly.
% Thus, 
\method uses a semi-supervised approach to expand the stylistic words from a handful number of human-annotated stylistics words. 
% in the benchmarking large style datasets without human annotations on lexica. 
First, we train \method using existing small human stylistic word annotation from \citet{hayati-etal-2021-bert}. Then, we obtain predicted stylistic words on the larger benchmarking style datasets and retrain \method on this expanded data to predict both the sentence's style label and the stylistic lexical explanation. 
% and automatically expand the lexicon set (Figure \ref{fig:task example} bottom). 
% \method explanations can also serve as a set of contextual stylistic lexica which has broader coverage compared to human-crafted lexicon dictionary (Figure \ref{fig:task example} bottom).

In this study, we show that for both in-domain and out-of-domain data, \method not only shows competitive classification performance with BERT-based model, but also generate stylistic lexical explanations that have higher alignment with human explanations. In terms of explanation quality, \method surpasses the baseline method across multiple explanation metrics. For the sufficiency metric, we improve upon the baseline by 14.12\% on the average. % with joy has the highest improvement (37.62\%). 
For plausibility, \method's lexical explanations correlate highly with human lexical annotation ground truth with an average Pearson's $r$ correlation score of 44\% compared to the baseline's correlation score of 3.9\%. 
% Moreover, we find that \method's lexical explanations appear more often in the hand-crafted stylistic lexical dictionary compared to the baseline's (60.26\% of \method important words vs. 38.78\% of baseline important words). 
Finally, we found that 72.5\% of \method's explanations are more preferred by human judges compared to the baseline.

% We find that \method's explanations sufficiently reflect human perception of stylistic words compared to lexical explanation baselines without sacrificing end-task accuracy on benchmarking style datasets we use for model training and out-of-domain datasets. In addition, we show that \method explanations are preferred by humans. 

% These results motivate future work on human-centered NLP, as we provide human perception-based explanations for correcting spurious behavior of NLP models and for better explaining linguistic styles. 

\begin{figure*}[t!]
    \centering
    \includegraphics[trim=0cm 9.5cm 0cm 0cm,clip,width=0.9\linewidth]{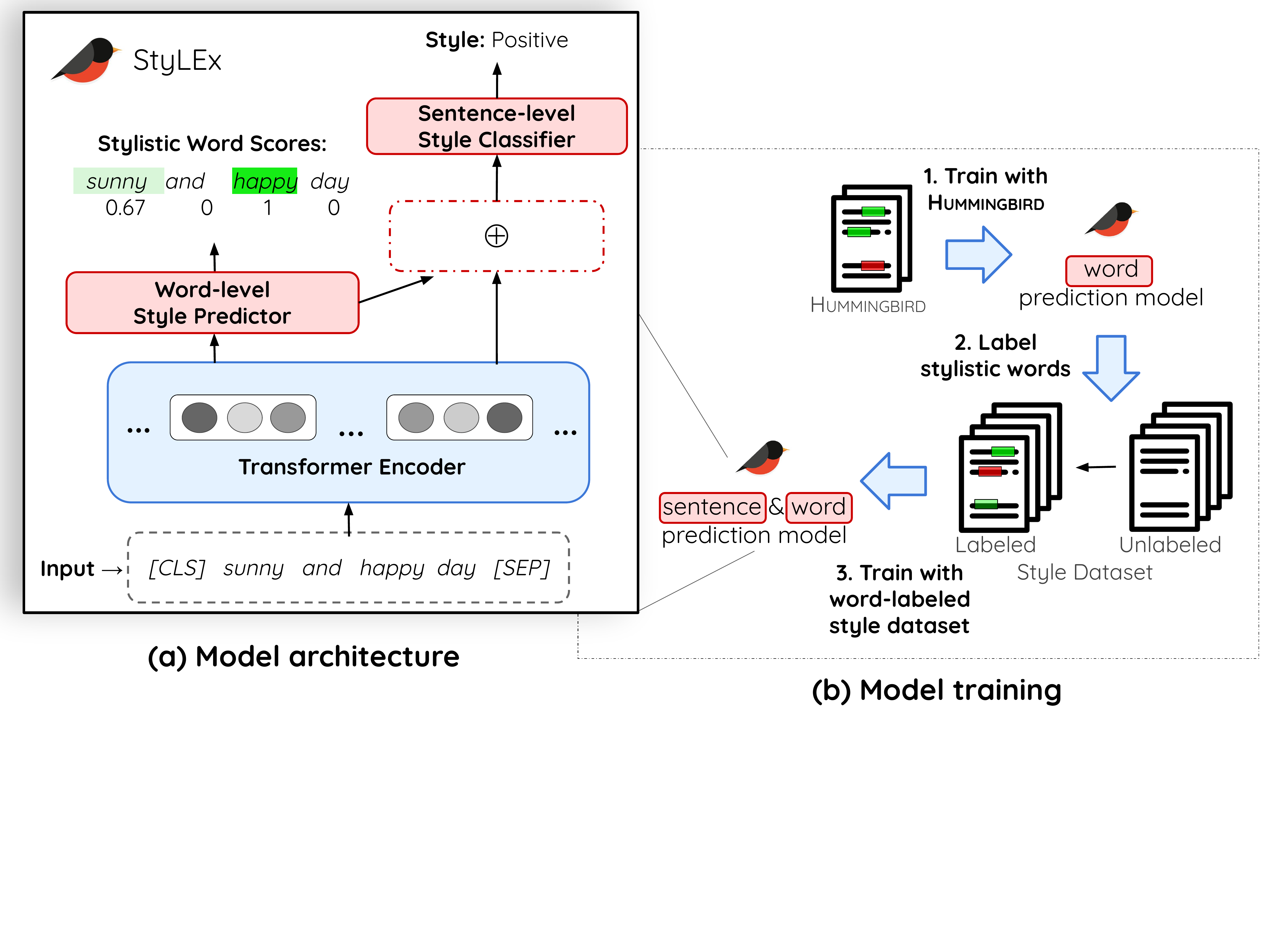}
    \caption{\textbf{(a)
    \method model architecture} (left). 
    Our model has two new modules: a \colorbox{darksalmon!40}{word-level} style predictor and a \colorbox{darksalmon!40}{sentence-level} style classifier. An aggregator appends the \colorbox{green!40}{word-level style logit} for each word to the hidden layer representations of each word and takes the max pooling of this aggregation.
    \textbf{(b) Model training} (right). 
    Human labels come from \textsc{Hummingbird} \cite{hayati-etal-2021-bert} for stylistic word scores and from \textsc{Original} datasets from sentence-level style classification. 
    (1) We train a stylistic \colorbox{darksalmon!40}{word-level} prediction model on \textsc{Hummingbird} dataset in order to (2) obtain \textit{pseudo-stylistic words} of sentences in the \textsc{Original} datasets. (3) Then we train another stylistic \colorbox{darksalmon!40}{word} and \colorbox{darksalmon!40}{sentence} prediction model on these \textsc{Original} sentences, now labeled with stylistic words.
    % \vspace{-2mm}
    }
    \label{fig:stylex_model}
\end{figure*}

\section{\method: Style Classification with Human Lexical Annotations}
% Given a style classification model that receives input sentence $\mathbf{x} = \{x_1, .. x_n\}$ where $x$ are words in the text and predicts the text's style label $y$, \method provides a set of lexical explanations $\mathbf{s} =\{s_1, ..., s_n\}$ that mimics human annotation of stylistic lexica where $s_i$ is the importance score of the $i$-th token $x_i$. 
% First, we explain \method architecture in \cref{stylex_section}. Then, we elaborate the data we use and \method model training in \cref{stylex_model_training}. 

% We explain how to incorporate lexicon-based human perception into \method architecture in . 
% More details of the training procedure are explained in \cref{model_training}. 

\subsection{\method Model Architecture}
\label{stylex_section}
\method is a joint model for word-level and sentence-level style prediction. Unlike a multi-task learning approach where tasks are independent of each other, \method exploits these stylistic word scores obtained from human annotation and then helps predict the sentence's styles. As displayed in Figure \ref{fig:stylex_model} (left), \method involves three modules: a transformer-based \cite{vaswani2017attention,devlin-etal-2019-bert} encoder, a word-level style predictor and a sentence-level style predictor. This work is based on BERT although the encoder can be applied to any transformer architecture.

Given an input of token sequence $\mathbf{x} = \{x_1, ..., x_n\}$ and its corresponding set of stylistic word scores $\{s_1, ..., s_n\}$, we encode $\mathbf{x}$ using a pretrained transformer model. We extract the final layer output as $\mathbf{h} = \{\mathbf{h}_1, ..., \mathbf{h}_n\}$ and feed $\mathbf{h}$ to the word-level style prediction layer which is a neural classifier that outputs stylistic word logits for each word $l_{{word}_{i}}$ computed as follows:
\vspace{-1mm}
\begin{equation*}
l_{{word}_{i}} = \mathbf{W}_{word}{\bf h}_i + {\bf b}_{word}
\end{equation*}

where $i \in \{1, .., n\}$, $\mathbf{W}_{word}$ is a matrix with the size ${H \times d_{l_{word}}}$, and $\mathbf{b}_{word} \in \mathbb{R}^{d_{l_{word}}}$ is the bias term. $H$ is the size of the default hidden layer in BERT which is 768 and $d_{l_{word}}$ denotes the number of classes of each style (e.g., positive or negative word in a sentiment classification task).

For the sentence-level style classification, we first take both the encoded representation $\mathbf{h}$ and stylistic word logits $l_{word}$. We then apply max pooling on the aggregation of $\mathbf{h} \oplus l_{word}$ along the sequence, resulting in vector $\mathbf{v} \in \mathbb{R}^{H + d_{l_{word}}}$ consisting of important logits. 
Finally, we input $\mathbf{v}$ into the sentence-level style classifier defined as follows:
% \vspace{-2mm}
\begin{equation*}
\begin{aligned}
l_{sentence} &= \texttt{softmax}(\mathbf{W}_{sentence}{\bf v} + {\bf b}_{sentence}) \\
P_{sentence} &= \argmax{(l_{sentence})}
\end{aligned}
\end{equation*}
where $l_{sentence} \in \mathbb{R} ^{2}$  denotes sentence-level style logits, $\mathbf{W}_{sentence}$ is a matrix with the size $(H + d_{l_{word}}) \times 2$, and $P_{sentence}$ is the index of the predicted sentence-level style. 

During training, \method's objective is to maximize the probability of the sentence's style and stylistic word scores.
The loss for both sentence-level style predictor and word-level style predictor is computed using binary cross entropy loss function. To jointly train the model, we optimize the following loss:
\vspace{-2mm}
\begin{equation*}
    \mathcal{L} = \mathcal{L}_{style} + \alpha \times \mathcal{L}_{word}
\end{equation*}
where $\alpha$ is a regularization hyperparameter. 

\subsection{\method Model Training}
To train \method, we need a dataset of stylistic sentences along with their corresponding stylistic words (\cref{styles_datasets}).  We use the \textsc{Hummingbird} dataset \cite{hayati-etal-2021-bert} that contains 500 sentences with word-level style annotation for obtaining the stylistic lexical explanation. Due to \textsc{Hummingbird}'s small size, we first train StyLEx on \textsc{Hummingbird} and then predict pseudo stylistic words on larger benchmark style datasets (> 6.8k sentences in the training sets) for training the final StyLEx model for both sentence classification and lexical explanation. 
\label{stylex_model_training}

\subsubsection{Datasets}
\label{styles_datasets}
Following \citet{hayati-etal-2021-bert}, we explore the same set of eight styles used in the dataset:  politeness, sentiment, offensiveness and five emotions (anger, disgust, fear, joy, and sadness) for style classification tasks. 
We use three sets of publicly available style datasets for our experiments as follows.

\paragraph{\textsc{Hummingbird}} is a multi-style dataset annotated with human perception scores on its important stylistic lexicons \cite{hayati-etal-2021-bert}. \textsc{Hummingbird} contains 500 sentences based on eight style datasets: politeness, sentiment, offensiveness, and five emotions (anger, disgust, fear, joy, and sadness). Three different crowd workers annotate each word in a sentence with 1 if they perceive the word as stylistic and 0 if not. The human perception score for a word is the average score of these annotators' labels. This perception score is what we call as stylistic word score and it is within the range [-1, 1]. We use \textsc{Hummingbird} for training \method's word-level style predictor. 

\paragraph{\textsc{Original} datasets} are used by \citet{hayati-etal-2021-bert} to curate \textsc{Hummingbird}.\footnote{We will refer to these individual datasets as ``\textsc{Original}''} Since some style labels in \textsc{Original} may contain continuous numbers rather than binary labels, we follow the same setting of \citet{hayati-etal-2021-bert} which only uses binary labels: polite or impolite, positive or negative, offensive or not offensive, anger or not anger, and so on. The politeness dataset comes from StackExchange and Wikipedia requests \cite{danescu-niculescu-mizil-etal-2013-computational} (9.8k training instances). The sentiment dataset is a collection of movie review texts \cite{socher-etal-2013-recursive} (117k training instances). The offensiveness dataset is from Twitter \cite{davidson2017automated} (20k training instances). The emotions dataset \cite{mohammad-etal-2018-semeval} is collected from tweets (6.8k training instances). For all these \textsc{Original} datasets, we use the default train/dev/split as explained in their papers. 

\paragraph{Out-of-Domain (\textsc{OOD}) datasets} are used to evaluate \method's performance on different domains. For each style, we use data from different sources or topics, but their style labels are in \textsc{Hummingbird} and \textsc{Original} datasets. For politeness, we use the polite and impolite sentences from the Enron email corpus \cite{klimt2004introducing, madaan-etal-2020-politeness}. For sentiment, we test \method on 5-core reviews from Amazon review dataset \cite{ni-etal-2019-justifying} for each product categories, except for movie reviews. We exclude movie reviews because it would be similar to the domain of the \textsc{Original}'s sentiment dataset. We convert ratings of 4-5 to positive labels and ratings of 1-2 as negative labels. For offensiveness, we use OffensEval \cite{zampieri2019semeval} dataset for offensiveness. For five emotions, we collect Reddit comments from GoEmotions corpus \cite{demszky2020goemotions}.\footnote{More details on the datasets are in Appendix \ref{sec:ood_data}.} 

\begin{table}[]
    \centering
    \setlength{\tabcolsep}{2pt}
    \begin{tabular}{l|c c|c  c}
        \toprule
        \multirow{2}{*}{\textbf{Style}} &
        \multicolumn{2}{c|}{\textbf{\textsc{Original} (\%)}}  & \multicolumn{2}{c}{\textbf{\textsc{OOD} (\%)}}
        \\ 
        & \textbf{BERT} & \textbf{\method} & \textbf{BERT} & \textbf{\method}
        \\
        \midrule
        Politeness & 67.96& 65.84 & 71.45 & 74.18 
        \\
        Sentiment & 96.52  & 96.59 & 86.70 & 86.99
        \\
        Offensiveness & 97.75 & 97.81 & 88.62 & 89.00
        \\
        Anger & 89.04 & 89.01 & 77.49 & 77.51
        \\
        Disgust & 86.50 & 86.90 & 74.06 & 74.63
        \\
        Fear  & 95.66 & 95.63 & 78.42 & 78.48
        \\
        Joy  & 88.02 & 88.12 & 75.20 & 74.26 
        \\
        Sadness & 88.38 & 88.41 & 78.37 & 78.71
        % \\ 
        % \hline
        % Average & \textbf{88.73} & 88.54 & 78.79 & \textbf{79.22}
        %& 71.34 & \textbf{85.47} 
        \\
        \bottomrule
    \end{tabular}
    \caption{For the sentence-level style classification task, \method does not sacrifice the task performance (F1-scores) of the \textbf{BERT} model across all of the style tasks across both \textbf{\textsc{Original}} and \textbf{\textsc{OOD}} settings.
    %*We have different ways of incorporating human perceptions in the Appendix, \method's joy result is the highest when we use another way of concatenation (see Table \ref{tab:ablation_results_emotion_offensive})
    % \vspace{-3mm}
    }
    \label{tab:experiment_result}
\end{table}

\subsubsection{Training}
\label{model_training}
The whole pipeline of \method model training is in Figure \ref{fig:stylex_model} (right). First, we train a stylistic word score prediction model with the same \method architecture in Figure \ref{fig:stylex_model} (left). We do this since the sentences in the benchmarking style datasets do not have human annotations of stylistic word scores. We then use a semi-supervised learning approach called, pseudo-labeling \cite{lee2013pseudo,rizve2020defense}, to label the stylistic words. Now the sentences in \textsc{Original} contain stylistic word scores which are output by the stylistic word predictor. Finally, we use both \textsc{Hummingbird} and \textsc{Original} for training another model of \method which predicts sentence-level binary style labels (polite and impolite, positive and negative etc.,) and provides lexical explanation scores within the range of [0, 1].\footnote{Other implementation details are in Appendix \ref{sec:implementation_details}.} 
% During training, \method’s objective is to maximize the probability of the sentence’s style and stylistic word scores with respect to the human-annotated labels: the sentence-level labels are from \textsc{Original} and word perception scores from \textsc{Hummingbird}.

\begin{table*}[]
    \centering
    % \small
    \begin{tabular}{@{}l l|l r@{}}
    \textbf{ID}& \textbf{Model} & \textbf{Sentence with Predicted Stylistic Word Scores} & \textbf{Sentence Style} \\ 
    \toprule
    \rowcolor{Gray} \multicolumn{4}{l}{Incorrect Baseline Prediction $\rightarrow$ Correct \method Prediction}\\ 
    \bottomrule
    \multirow{2}{*}{\textsc{Orig}-1} & \method & {\fontfamily{lmss}\selectfont ... because i'm gonna ' add \colorbox{mygreen}{insult} to \colorbox{palegreen}{injury} } & Disgust \checkmarknew\\
     & Baseline& {\fontfamily{lmss}\selectfont ... because i'm gonna ' add \colorbox{lightsalmon}{insult} \colorbox{palegreen}{to} \colorbox{darksalmon}{injury} } & Not Disgust \crossmark[red]\\
    \hline
    \textsc{Orig}-2 & \method & {\fontfamily{lmss}\selectfont ... yet you say i'm a \colorbox{mygreen}{stogie} you're your own \colorbox{mygreen}{downfall}} & Offensive \checkmarknew\\ 
    & Baseline & {\fontfamily{lmss}\selectfont ... yet you say i'm a \colorbox{palegreen}{stogie} you're your own \colorbox{lightsalmon}{downfall}} & Not Offensive \crossmark[red] \\ 
     \toprule
     \rowcolor{Gray} \multicolumn{4}{l}{Correct Baseline Prediction $\rightarrow$ Incorrect \method Prediction}\\ \bottomrule
     \multirow{2}{*}{\textsc{Orig}-3} & \method & {\fontfamily{lmss}\selectfont \colorbox{mygreen}{please} put them all back are you on dsl} & Polite \crossmark[red]\\
    & Baseline & {\fontfamily{lmss}\selectfont \colorbox{mygreen}{please} \colorbox{darksalmon}{put} \colorbox{lightsalmon}{them} all \colorbox{darksalmon}{back} \colorbox{palegreen}{are} \colorbox{lightsalmon}{you on} \colorbox{mygreen}{ds}l} & Impolite \checkmarknew\\ \hline
    \multirow{4}{*}{\textsc{Orig}-4} &  \multirow{2}{*}{\method} & {\fontfamily{lmss}\selectfont ... can't just be mean and do \colorbox{palestgreen}{horrid things busy without}} & \multirow{2}{*}{Not Fear \crossmark[red]}\\
     &  & {\fontfamily{lmss}\selectfont paying the price} & \\
     &  \multirow{2}{*}{Baseline} & {\fontfamily{lmss}\selectfont ... can't just be mean and do \colorbox{mygreen}{horrid} \colorbox{palegreen}{things} busy \colorbox{palegreen}{without} } & \multirow{2}{*}{Fear \checkmarknew}\\
     &  & {\fontfamily{lmss}\selectfont paying the \colorbox{palestgreen}{price}} & \\
    \bottomrule
    %  \multicolumn{4}{l}{\textbf{\textsc{OOD}: Incorrect Baseline Prediction $\rightarrow$ Correct \method Prediction}}\\ \hline
    % \textsc{OOD}-5 & \method & {\fontfamily{lmss}\selectfont \colorbox{lightsalmon}{to be} \colorbox{lightblue}{updated}} & Polite  \checkmarknew\\
    % \textsc{OOD}-6 & \method & {\fontfamily{lmss}\selectfont \colorbox{lightsalmon}{to be} \colorbox{lightblue}{updated}} & Positive  \checkmarknew\\
    % \textsc{OOD}-7 & \method & {\fontfamily{lmss}\selectfont \colorbox{lightsalmon}{to be} \colorbox{lightblue}{updated}} & Positive  \checkmarknew\\
    \end{tabular}
    \caption{Error analysis on prediction-flipped sentences. Baseline refers to sentence-level style prediction result by fine-tuned BERT model and highlights are stylistic words found by the integrated gradient method. Green highlights on the words mean that the model predicts high positive word-level stylistic scores; red for the opposite label (e.g., \textit{impolite} or \textit{negative}). Sentence Style is a model's sentence-level style prediction \checkmarknew marks correct prediction and \crossmark[red] denotes incorrect prediction. 
    % \shirley{Add more samples if time permits, add more failure cases in the limitation e.g. highlight tokens are sparse, explain potential future work}
    \vspace{-3mm}}
    \label{tab:example_classified_sentences}
\end{table*}

\section{Evaluation on Style Classification}
\subsection{Baseline}
To assess \method's classification performance, we compare it with a fine-tuned BERT-based classifier as a baseline. The training data for the baseline is also a combination of \textsc{Hummingbird} and \textsc{Original}. For explanation evaluation, we compare \method's explanation with the commonly-used explanation method called integrated gradients \cite{mudrakarta-etal-2018-model, sundararajan2017axiomatic}, implemented in Captum\footnote{\url{https://captum.ai}}. Integrated gradient, which can be viewed as an approximate method of estimating Shapley values, is defined as follows. For the input sequence of words $x$ and a neural network function $F$, an attribution score (the explanation) for each word is defined as the gradient between the input $x$ and baseline $x'$ of the function $F$ where $x'$ is a zero scalar. 

% \subsection{\method Implementation Details}
% Throughout the experiment, we set $d_{l_{word}} = 2$ for politeness (polite, impolite) and sentiment (positive, negative) which have two perception classes and $d_{l_{word}} = 1$ for the other styles. At the loss calculation step, we set the regularization hyperparameter $\alpha$ to 0.05 which gives the best style and perception prediction found searching the range [0.01, 100]. 
% For the pseudo-labeling approach, we use the same architecture and hyperparameters with \method model. We first train \method with \textsc{Hummingbird} training set only to predict perception scores for 50 epochs. Then we select the model with best F1 score as a perception predictor to provide pseudo perception labels for tokens in \textsc{Original} training set. Then, we use both human-annotated perception score from \textsc{Hummingbird} and pseudo perception scores from \textsc{Original} to train the sentence-level style prediction as in Figure \ref{fig:stylex_model}. For the sentence-level model, we train the model for 5 epochs. For both perception prediction and sentence style classification, we use BERT-base-uncased pretrained model. We set 0.1 dropout rate, 512 maximum sequence length, AdamW optimizer of learning rate $2e^{-5}$. For other hyper-parameters, we follow the default setting from HuggingFace’s transformer library \cite{wolf-etal-2020-transformers}.

\subsection{Results}
In our experiment, we have eight \method models for each style: politeness, sentiment, offensiveness, and five emotions. For each style, we run \method on the \textsc{Original} test sets and \text{OOD} datasets for five times with different seeds and report the average of F1-scores in Table \ref{tab:experiment_result}. 
For \textsc{Original} datasets, \method achieves higher F1 scores compared to the fine-tuned BERT model on sentiment, offensiveness, disgust, joy, and sadness. 
Overall, we observe that \method does not sacrifice task performance of the state-of-the-art classifiers while predicting stylistic word scores.
When tested on the \textsc{OOD} test sets, \method achieves higher F1 score against the fine-tuned BERT model for all styles. Politeness has the greatest improvement from 71.48\% to 74.18\% since we observe that the \textsc{Original} dataset of politeness contains many spurious content words. When we use bigger language models such as RoBERTa \cite{liu2019roberta}, XLNet \cite{yang2019xlnet}, and T5 \cite{raffel2020exploring} for \method, \method still has better results than the baseline for five styles: sentiment, offensiveness, anger, and disgust.

From example sentences from the test sets in Table \ref{tab:example_classified_sentences}, we can see how \method helps task performance. \textsc{Orig}-1 and \textsc{Orig}-2 sentences show how \method can capture stylistic words and correct the sentence's style label to \textit{disgust}. For example, ``insult'' and ``injury'' in \textsc{Orig}-1 are initially labeled by the integrated gradient method as unimportant for identifying \textit{disgust}, but \method identifies the words as stylistic cues. Similarly, for the word ``downfall'' in \textsc{Orig}-2, \method finds it as offensive, but the baseline does not. \method also has a higher stylistic word score for indicating ``stogie'' as an offensive word. 

As we look at the politeness classification results, we find that most of the incorrect cases are when \method mislabels subtle \textit{impolite} sentences as \textit{polite}. As we observe in Table \ref{tab:example_classified_sentences} \textsc{Orig}-3, \method finds the word ``please'' as a polite cue, but the ground truth label of the sentence is \textit{impolite}. We then inspect its continuous score from the original politeness dataset by \citet{danescu-niculescu-mizil-etal-2013-computational}. It turns out that its politeness score is $-0.38$ in the range of [-2, 2] as -2 being the most impolite sentence. This shows that this sentence score is closer to neutral than impolite. 
% Unfortunately having more number of labels may also lower the overall performance. 
This finding also reflects how \textsc{Hummingbird} dataset has been collected: as mentioned in \citet{hayati-etal-2021-bert}, words from the offensive dataset (mostly swear words) are often labeled as impolite by human annotators. Thus, it may bias the annotators' view that sentences with impolite labels are not as bad as offensive sentences, making them not mark offensive sentences as \textit{impolite}. Therefore, for such subtle impolite sentences, human annotators in \textsc{Hummingbird} may not label the sentence and words as impolite.

In contrast, \method misclassifies an \textit{impolite} sentence as \textit{polite} and \textit{fear} as \textit{not fear}. As we look at \textsc{Orig}-4 in Table \ref{tab:example_classified_sentences}, \method weakly finds \textit{fear} cues (``horrid'', ``things'') but they do not help in boosting the model to predict the sentence as \textit{fear}. We conjecture that this is because there are very few training samples labeled with \textit{fear} and \textit{fear} has quite low word-level inter-annotator agreement as reported in \citet{hayati-etal-2021-bert}.

\section{Style Explanation Evaluation}
We investigate \method's explanations if they are sufficient, plausible, and understandable following previous works \cite{deyoung-etal-2020-eraser, jacovi-goldberg-2020-towards, wiegreffe2021teach,rajagopal2021selfexplain}. 
% In order to evaluate whether \method's explanations are faithful, we run sufficiency test. 
\citet{jacovi-goldberg-2020-towards} define that a faithful interpretation represents a model's reasoning process. To evaluate whether \method's explanations are faithful, we run a sufficiency test that evaluates whether the model explanations alone are highly relevant for predicting the label \cite{jacovi-etal-2018-understanding}. 
Meanwhile, we measure plausibility to examine whether the explanation is agreeable to humans \cite{deyoung-etal-2020-eraser}. 
Finally, understandability measures if a user is able to understand model explanations \cite{rajagopal2021selfexplain}. 
For investigating sufficiency and plausibility, we run automatic metrics. 
To assess understandability, we ask human judges to choose the explanation that can be better understood by a non-expert between \method and the integrated gradients (IG) method. 

\begin{table}[]
    % \small
    \centering
    \setlength{\tabcolsep}{3pt}
    \begin{tabular}{l|c c| c c}
        \toprule
        \multirow{2}{*}{\textbf{Style}} & \multicolumn{2}{c|}{\textbf{\textsc{Original}}} & \multicolumn{2}{c}{\textbf{\textsc{OOD}}}\\
        & \textbf{IG} & \textbf{\method} & \textbf{IG} & \textbf{\method}
        \\
        \midrule
        Politeness & 43.92 &  \textbf{63.08} & 52.89 & \textbf{68.19}
        \\
        Sentiment & 87.18 & \textbf{89.39} & 64.93 & \textbf{77.52}
        \\
        Offensiveness & 84.87 & \textbf{91.26} & 82.93 & \textbf{84.75}
        \\
        Anger & 68.36 & \textbf{86.90} & 53.76 & \textbf{73.99}
        \\
        Disgust & 82.54 & \textbf{85.91} & 71.36 & \textbf{75.76}
        \\
        Fear & 87.82 & \textbf{96.10} & 35.85 & \textbf{65.26}
        \\
        Joy  & 45.54 & \textbf{83.16} & 55.49 & \textbf{72.71}
        \\
        Sadness &  70.49 & \textbf{87.94} & 47.83 & \textbf{70.95}
        \\
        \bottomrule
    \end{tabular}
    \caption{The results for sufficiency test on the \textbf{\textsc{Original}} and \textbf{\textsc{OOD}} data show the F1 scores on top-$k$ words. IG stands for integrated gradient.
    }
    \label{tab:sufficiency_test_results}
\end{table}
\subsection{Sufficiency}
Following \citet{jain2020learning,rajagopal2021selfexplain}'s sufficiency test, we fine-tune a BERT model on the top-$k$ words as explanation instead of the whole sentence. We limit an explanation to contain 30\% words of the average sentence length for each of the style datasets. These words are ranked based on their importance score by the baseline integrated gradient method and \method for all the positive stylistic words (polite words, positive words, offensive words, angry words). 

In Table \ref{tab:sufficiency_test_results}, we can see that explanations from \method show much higher predictive performance compared to explanations extracted by the integrated gradient method for all styles in both datasets, \textsc{Original} and \textsc{OOD}. This result suggests that human-like stylistic words are much more strongly predictive of a sentence's style compared to the gradient-based explanation methods that often rely on content words as an explanation. This indicates that \method's explanations are relatively more faithful compared to the integrated gradients based method.

\subsection{Plausibility}
We use two approaches to measure the agreement between \method's lexical explanations and stylistic words perceived by humans to assess the plausibility of \method's explanations. In the first approach, we compare \method's stylistic words on the \textsc{Hummingbird} test set and compare it with the ground truth human perception scores in \textsc{Hummingbird}. Second, we compare \method's top-$k$ stylistic words with existing expert-curated stylistic lexicon dictionaries. The domain of the existing expert-curated stylistic lexicon dictionaries could be different from the domain of the datasets in our study which range from social media texts to Wikipedia. However, it is still useful to compare how much \method and the baseline agree human experts on identifying stylistic words. 

\begin{figure}
    \centering
    \includegraphics[trim=0cm 0.1cm 0.8cm 0cm,clip,width=0.9\linewidth]{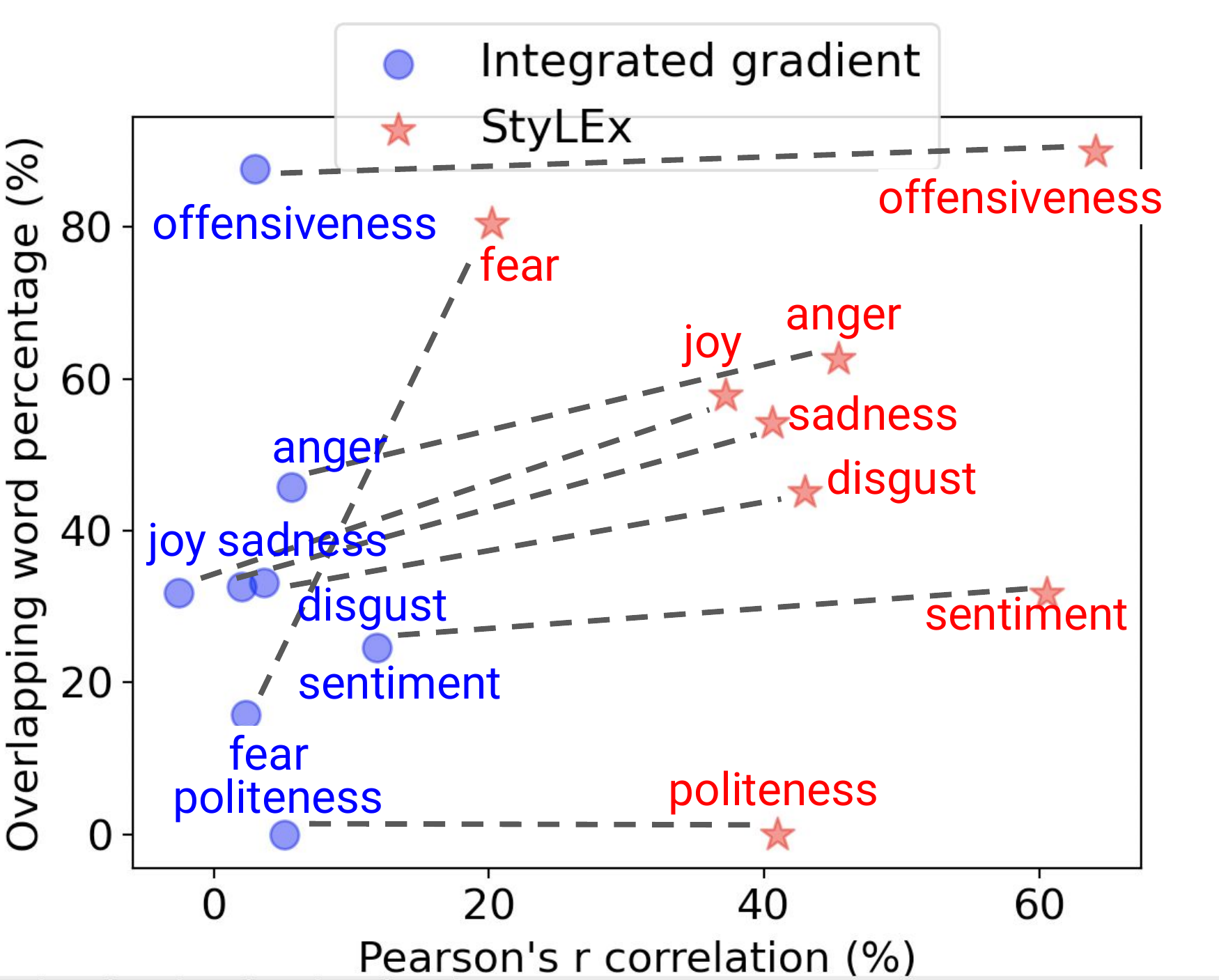}
    \caption{Plausibility experiment result. There are two points for each style in this plot. A blue circle point is for the baseline IG method and a red star point for \method. X-axis is Pearson's $r$ correlation score for each style. Y-axis is the percentage of stylistic sentences with style words appearing in the existing style lexicon dictionary.}
    % \vspace{-4mm}}
    \label{fig:plausibility}
\end{figure}

Figure \ref{fig:plausibility} shows a scatterplot of \method vs. the integrated gradient. X-axis represents the Pearson's $r$ correlation score on the \textsc{Hummingbird} test set.
Y-axis is the percentage of overlapping words between the important words found by \method and the baseline compared with the human-curated style lexicon dictionary. We calculate the overlapping word percentage as follows. We compute how many of the top 30\% of the stylistic words in the \textsc{Original} datasets found by \method or baseline appear in human-curated dictionaries for the emotion/sentiment/offensive lexicons.

In Figure \ref{fig:plausibility}, the higher the Pearson’s correlation score is (to the right), the better the explanation words produced by the model (\method or baseline) are aligned with human perception ground truth from \textsc{Hummingbird}. The dashed lines show how much StyLEx’s generated stylistic words align more with human annotations for stylistic words from both \textsc{Hummingbird} and human-curated stylistic lexicon dictionaries. 

\begin{table*}[]
    \centering
    \begin{tabular}{l |l | l | l}
    \toprule
        \textbf{Style} &\raisebox{-2pt}{\includegraphics[height=1em]{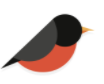}}\raisebox{-2pt}{\includegraphics[height=1em]{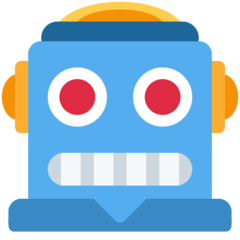}} \textbf{Both}
        & \raisebox{-2pt}{\includegraphics[height=1em]{emoji/bird.png}}\textbf{\method}
        & \raisebox{-2pt}{\includegraphics[height=1em]{emoji/robot.png}} \textbf{Integrated gradient}
        \\
        \midrule
        \textbf{Positive} &  good, fun, love & \textcolor{red}{associate, develop, instruct}  &  \textcolor{blue}{deserve, endure, football}
        \\
        \textbf{Negative} & bad, horror, silly  & \textcolor{red}{mess, chaos, disappoint} & \textcolor{blue}{maternal, banger, yell}
        \\
        \textbf{Offensive} & bitch, bitches, pussy & \textcolor{red}{blind, racist, panties} & \textcolor{blue}{fairy, amateur, fisting} \\
        % & hoes, ass & & \textcolor{red}{turd, milf} & & \textcolor{blue}{bible, condom}
        % \\
        \textbf{Anger} & angry, anger, awful & \textcolor{red}{frowning, scare, lose} & \textcolor{blue}{belt, campaigning, destroying}*\\
        % & & &
        % \\
       \textbf{Disgust} & awful, terrible, angry  &\textcolor{red}{dismal, frowning, animosity} & \textcolor{blue}{congress, finally, sentence}* \\
    %   & & &
    %     \\
        \textbf{Fear} & fear, anxiety, nervous &  \textcolor{red}{horrid, war, threaten} &  \textcolor{blue}{rejects, mum, beating} \\
        % & & &
        % \\
        \textbf{Joy} & happy, love, good &  \textcolor{red}{faith, sing, succeed} &  \textcolor{blue}{deal, independence, football}
        % & & &
        \\
        \textbf{Sadness} &  depression, sadness, lost & \textcolor{red}{bad, offended, leave} & \textcolor{blue}{funeral,  bloody, case}*\\
        \bottomrule
    \end{tabular}
    \caption{Three important words found by \method (\raisebox{-2pt}{\includegraphics[height=1em]{emoji/bird.png}}) and the integrated gradient method (\raisebox{-2pt}{\includegraphics[height=1em]{emoji/robot.png}}) that appear in the stylistic lexicon dictionary.* = words only appear one time in the test data.}
    \label{tab:stylistic_words}
\end{table*}

\paragraph {(1) Correlation with human perception.} We investigate how similar \method's explanations are with human perceptions. To do so, we compute the Pearson's correlation $r$ between stylistic word scores predicted by StyLEx and annotated by humans from \textsc{Hummingbird} annotations for each word by concatenating all the predicted stylistic word scores. Our correlation score is different from \citet{hayati-etal-2021-bert} because we split the Hummingbird dataset into a training set and test set even though we use the same human perception scores and the integrated gradient method. The correlation score we reported in Figure \ref{fig:plausibility} is from the \textsc{Hummingbird} test set.

In Figure \ref{fig:plausibility} (vertical trend), we can see that \method explanations correlate more with ground truth human perception for all styles, as red stars are stretched to the right. Sentiment and offensiveness are styles that have the highest correlation scores (60.53\% and 64.09\%) while fear is the lowest (20.17\%)). Explanations from integrated gradient correlate very loosely with human perception ground truth with sentiment as the highest (11.89\%) and joy negatively correlates with human perceptions (-2.55\%). 

\paragraph{(2) Comparison with stylistic lexicon dictionaries.}
We then investigate how similar the stylistic words found by \method are to the stylistic words curated by humans in the existing lexicon dictionary. We use sentiment emotion lexicons from \citet{mohammad-turney-2010-emotions} and offensive lexicons from \citet{luis2021}.\footnote{We couldn't find a publicly available politeness lexicon dictionary.}
Using the same set up of sufficiency test, we select top-30\% stylistic words from each sentence in \textsc{Original} datasets with the positive style label. Then we check if at least one of these words appear in the existing lexicon dictionary and compute its average across all training samples.

In Figure \ref{fig:plausibility} (horizontal trend), we see that \method consistently has higher percentage of word occurrences in the lexicon dictionary compared to the integrated gradient method where fear has the highest percentage difference (15.78\% $\rightarrow$ 80.43\%) and offensiveness has the lowest percentage change (87.67\% $\rightarrow$ 89.99\%). Averaging across all styles, we find that 56.70\% of the stylistic sentences with \method stylistic words appear in the existing style lexicon dictionary while integrated gradient only identifies 37.01\% of those words.

The score is higher for offensiveness than for sentiment or emotion. We observe that people use more offensive words in social media, which is the source for dataset collection. We also examine  the lower occurrence for emotions. From our analysis, we found that the emotion lexicon dictionary contains several colloquially rare words ``aberration'' or ``meritorious'', leading to a very low overlap with the datasets that we used for the analysis. 

We also take a closer look at how many and the nature of important words are captured by \method and/or the integrated gradient method as shown in Table \ref{tab:stylistic_words}. These word scores are obtained by averaging their scores and then we sort them based on these average scores. In general, we find that \method can find more diverse stylistic words as defined in the existing lexicon dictionary for all styles except for positive sentiment. Some emotion words found by the integrated gradients only appear rarely in the data (mostly only once). 

\begin{figure}
    \centering
    \includegraphics[trim=0.2cm 0.2cm 0.2cm 0.2cm,clip,width=\linewidth]{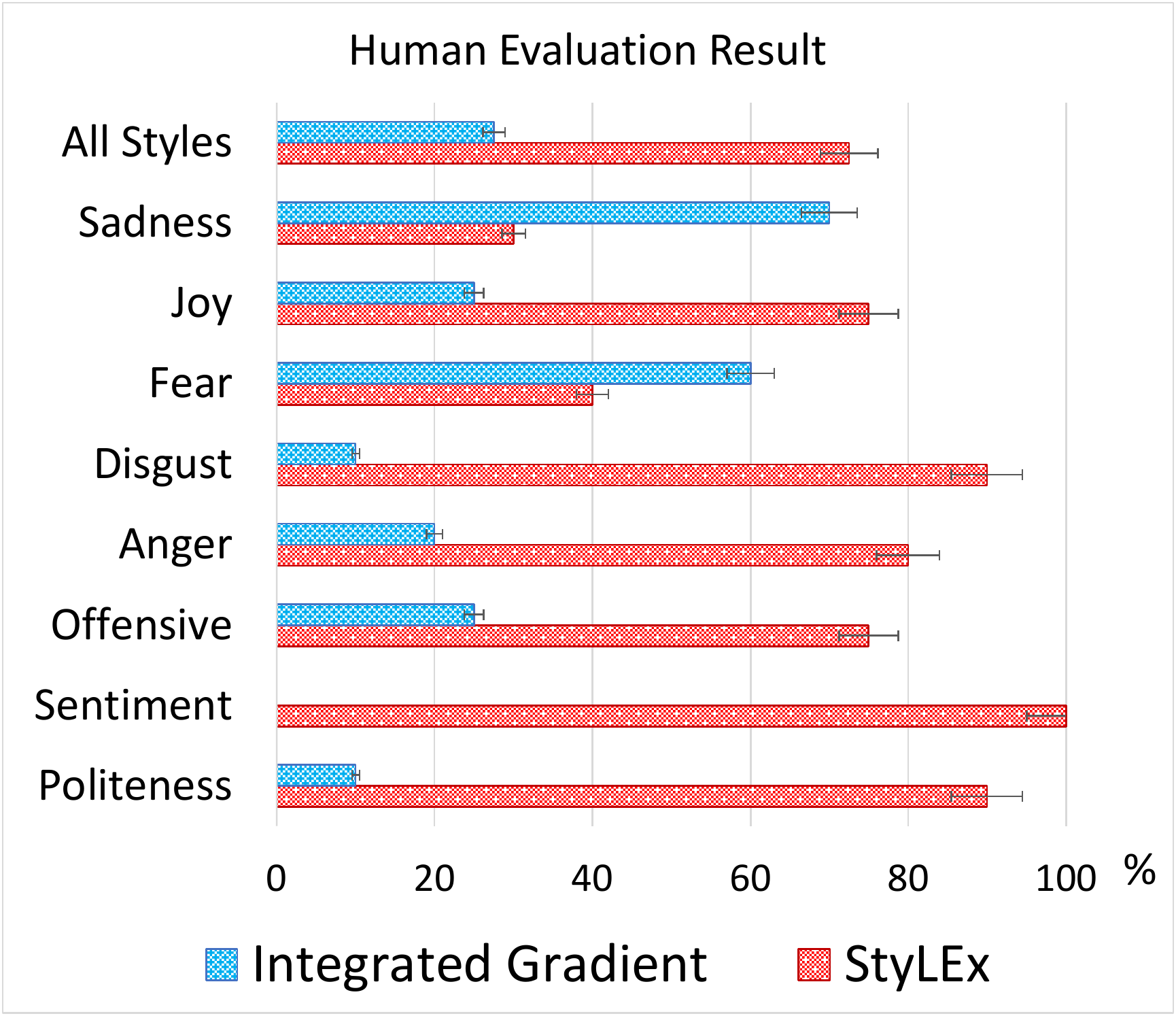}
    \caption{Human evaluation results. X-axis is the percentage of explanations preferred by human judges.
    \vspace{-3mm}}
    \label{fig:human_eval_figure}
\end{figure}

\subsection{Understandability}
To investigate the quality of \method's explanation, we ask human judges to evaluate \method's explanations compared to baseline explanations. {\it Understandability} asks whether human judges understand our explanation better than the explanation computed using integrated gradients. For this study, we randomly select 20 stylistic sentences for each of the eight styles, resulting in total 160 sentences. These 20 sentences are constructed by 10 sentences from \textsc{Original} test set and 10 sentences from \textsc{OOD} test set. We normalize the stylistic word scores for sentence length across all sentences. 
% For both \method and the integrated gradient method, we restrict to selecting a sentence which has at least perception score/importance score > 0.25 to make meaningful highlights \footnote{some weak sentences can have very pale highlights, leading to inconclusive results}. 

A human judge is shown two versions of the same sentence with different anonymized highlights, as shown in Figure \ref{fig:task example}. We then ask three different human judges to select (through Amazon Mechanical Turk) the explanation that was more understandable. Each worker annotated 20 sentences of the same style. The order of explanations is randomized to remove bias. We say that an explanation by a method is preferred by human judges when the majority choose that method. If the majority chooses a method for all the 20 sentences, the X-axis score will be 100\%. Results in Figure \ref{fig:human_eval_figure} show that across all styles (160 pairs of explanations), \method gives an overall gain of 27.5\%. 

% Another reason would be that the lack of annotated lexicons from \textsc{Hummingbird} leads to less diverse set of pseudo-labels to be highlighted in \method training.

% Despite the small data size of human perception used in our training, 

% \paragraph{\method vs. Hummingbird}
% \method's output is sparse, distribution of how many lexicon explanations are produced by \method vs. human (what if they are similar distribution?

\section{Related Work}
% \begin{table*}[]
%     \centering
%     \begin{tabular}{l|l l l}
%          & Lexicon-based & Model explanation & \method\\ 
%          & Classification & \\
%          \hline
%         Prior work &  & \\
%         Covered styles & Sentiment & \\
%         Pros \& cons & & \\
%     \end{tabular}
%     \caption{Comparison}
%     \label{tab:related_work}
% \end{table*}
\paragraph{Styles in NLP} Research on style in NLP has addressed various tasks including style classification \cite{danescu-niculescu-mizil-etal-2013-computational, socher-etal-2013-recursive}, style transfer \cite{rao2018dear,li2018delete}, style and content disentanglement \cite{john2018disentangled,zhu2021neural}, and multiple style analysis \cite{hayati-etal-2021-bert, kang-hovy-2021-style}. This work focuses on understanding stylistic variation in style classification.
%, since it is cleaner to analyze than stylistic text generation.
Style classification models often produce spurious features \cite{sen-etal-2021-counterfactually,schlangen-2021-targeting, bras-v119-bras20a}, motivating us to leverage stylistic variation from human perspectives to distinguish between stylistic words and content words. Past work have used stylistic lexica for classification \cite{taboada2011lexicon, eisenstein2017unsupervised}, but in this work, we fine-tune the language model to generate these lexica and use them to help style prediction. Our work is most closely related to \citet{hayati-etal-2021-bert}, but they do not develop any new models to use the human perception scores as explanations. Moreover, while linguistics styles can cover an author's writing style or figurative language, we limit our study to high-level style as used in \citet{kang-hovy-2021-style, hayati-etal-2021-bert}.

\paragraph{Explainable NLP}
Heat maps generated from attention values from the models \citep{bahdanau2014neural} are widely used as an interpretability tool, but these attention maps are often unfaithful and unreliable  \citep{jain-wallace-2019-attention,wiegreffe-pinter-2019-attention,zhong2019fine,pruthi2019learning}. Saliency maps computed via gradients offer an alternative \citep{sundararajan2017axiomatic,smilkov2017smoothgrad,mudrakarta-etal-2018-model}. 
Annotating explanations as \emph{rationales} (part of input) \citep{lei-etal-2016-rationalizing} through expert annotations \citep{zaidan2008modeling} is widely used to model explanations in NLP when external annotations are available. Another class of inherently interpretable models aims to optimize model explanations without any external annotations \cite{card2019deep,croce2019auditing,rajagopal2021selfexplain}. Our work is similar in spirit to the rationale approaches \citep{lei-etal-2016-rationalizing} but focuses on understanding style attributes in text and computationally modeling them based on human annotation of the important words.

\section{Conclusion}
We proposed \method, a style classification model for learning stylistic variations through lexical explanation. With only 500 sentences with word-level style annotation, we find improvement in both classification and explanation. Compared to the commonly-used integrated gradient method, \method's explanations are more accurate for model prediction, more consistent with human-found stylistic words from existing datasets and lexicon dictionaries, and better understandable by human judges, without sacrificing task performance on both in-domain and out-of-domain datasets.

% Moreover, we find that \method is conservative in selecting important words in sentences, leading to fewer words being highlighted, despite the high precision of the highlighted words.
% Compared to human perceptions annotated in \textsc{Hummingbird} and model's perceptions using integrated gradients, \method has only XX.X\% and YY.Y\% of stylistic words predicted, respectively.
% We conjecture that this is in part because our model uses the perception annotations only when more than two annotators out of three agreed.

\paragraph{Future Work} Our approach opens up future work on human-centered lexical explanation for correcting the spurious behavior of NLP models and for better explaining linguistic styles. We plan to investigate collecting more human lexical annotations to more accurately model stylistic variation, especially with larger pretrained language models. Broader usage of \method in providing stylistic cues will be applicable to lexical style and content disentanglement \cite{cheng2020improving, john2019disentangled}, counterfactual data augmentation for style-related tasks \cite{sen-etal-2021-counterfactually}, and stylistic paraphrasing \cite{pavlick-nenkova-2015-inducing}. 

\paragraph{Limitations} Our work has some limitations, mostly stemming from the size and nature of the human-annotated data. The training data (500 sentences) from \textsc{Hummingbird} is quite small to train deep learning models. However, our work shows that with just 500 sentences we could achieve a huge improvement in interpretability as well as a slight improvement in OOD performance, using semi-supervised training. Moreover, there is sparsity of stylistic words in the sentences. We also found that some stylistic words have various scores of human perception; capturing such subtle stylistic words is difficult. An interesting future work would be to handle these problems of sparsity and subtlety. We also notice that \textsc{Hummingbird} is annotated by people residing in the United States. Thus, their perception of styles may not reflect the perception of those with different cultural backgrounds. 
%Nevertheless, \method is applicable to any kind of dataset training with similar human lexical annotation; \method is not limited to being used on the \textsc{Hummingbird} dataset. 
Nevertheless, \method can be applied to any dataset training with similar human lexical annotations, and not limited to \textsc{Hummingbird}. 

\paragraph{Ethical Considerations} When collecting the explanation evaluation from human judges, we inform them that the content may contain offensive languages that could be upsetting. 

\section*{Acknowledgments}
We would like to thank Karin de Langis for her valuable feedback during the early version of writing and the anonymous reviewers for their thoughtful comments.

\bibliography{custom}
\bibliographystyle{acl_natbib}

\appendix

\section{Appendix}
\label{sec:appendix}

\subsection{Sampling \textsc{OoD} Data and Data Statistics}
\label{sec:ood_data}
\begin{itemize}
% [noitemsep,topsep=0pt,leftmargin=*]
    \item Politeness: We randomly sample 500 polite sentences and 500 impolite sentences from the Enron email corpus \cite{klimt2004introducing, madaan-etal-2020-politeness} since the size of entire corpus (>600k) is too large for inference. 
    
    \item Sentiment: We test \method on 5-core reviews from Amazon review dataset \cite{ni-etal-2019-justifying}. For each category, we sample 100 positive sentences and 100 negative sentences from review categories, except for movie reviews which would be similar to the domain of the \textsc{Original} dataset. We convert ratings of 4-5 to positive labels and ratings of 1-2 as negative labels. 
    
    \item Offensiveness: We use OffensEval \cite{zampieri2019semeval} dataset for offensiveness. We select all offensive tweets (3,002 instances) and all non-offensive tweets (2,991 instances) since OffensEval dataset is already nearly balanced.
    
    % \textcolor{blue}{We use all negative \sh{Kyumin, is this offensive label or non-offensive label?)} data and randomly undersample positive data to match the number with negative samples.}
    
    \item Emotions: For five emotions, we collect samples from GoEmotions corpus \cite{demszky2020goemotions} that contains Reddit comments labeled with 27 emotions, but we only select the five relevant emotions. For each emotion, we use all data for the positive emotion (e.g., joy) and undersample the negative emotion (e.g., not joy) data to equal the number of positive emotion samples.
\end{itemize}

The three dataset statistics are summarized in Table \ref{tab:data_stats}. 

\subsection{\method Implementation Details}
\label{sec:implementation_details}
Throughout the experiment, we set $d_{l_{word}} = 2$ for politeness (polite, impolite) and sentiment (positive, negative) which have two style classes and $d_{l_{word}} = 1$ for the other styles. At the loss calculation step, we set the regularization hyperparameter $\alpha$ to 0.05 which gives the best style and perception prediction found searching the range [0.01, 100]. 
For the pseudo-labeling approach, we use the same architecture and hyperparameters with \method model. We first train \method with \textsc{Hummingbird} training set only to predict stylistic word scores for 50 epochs. Then we select the model with the best F1 score as a stylistic word score prediction to provide stylistic word scores for tokens in \textsc{Original} training set. Then, we use both human-annotated perception score from \textsc{Hummingbird} and predicted stylistic word scores from \textsc{Original} to train the sentence-level style prediction as in Figure \ref{fig:stylex_model}. For the sentence-level model, we train the model for 5 epochs. For both stylistic word prediction and sentence-level style classification, we use BERT-base-uncased pretrained model. We set 0.1 dropout rate, 512 maximum sequence length, AdamW optimizer of learning rate $2e^{-5}$. For other hyper-parameters, we follow the default setting from HuggingFace’s transformer library \cite{wolf-etal-2020-transformers}.

\begin{figure*}
    \centering
    \includegraphics[width=\linewidth]{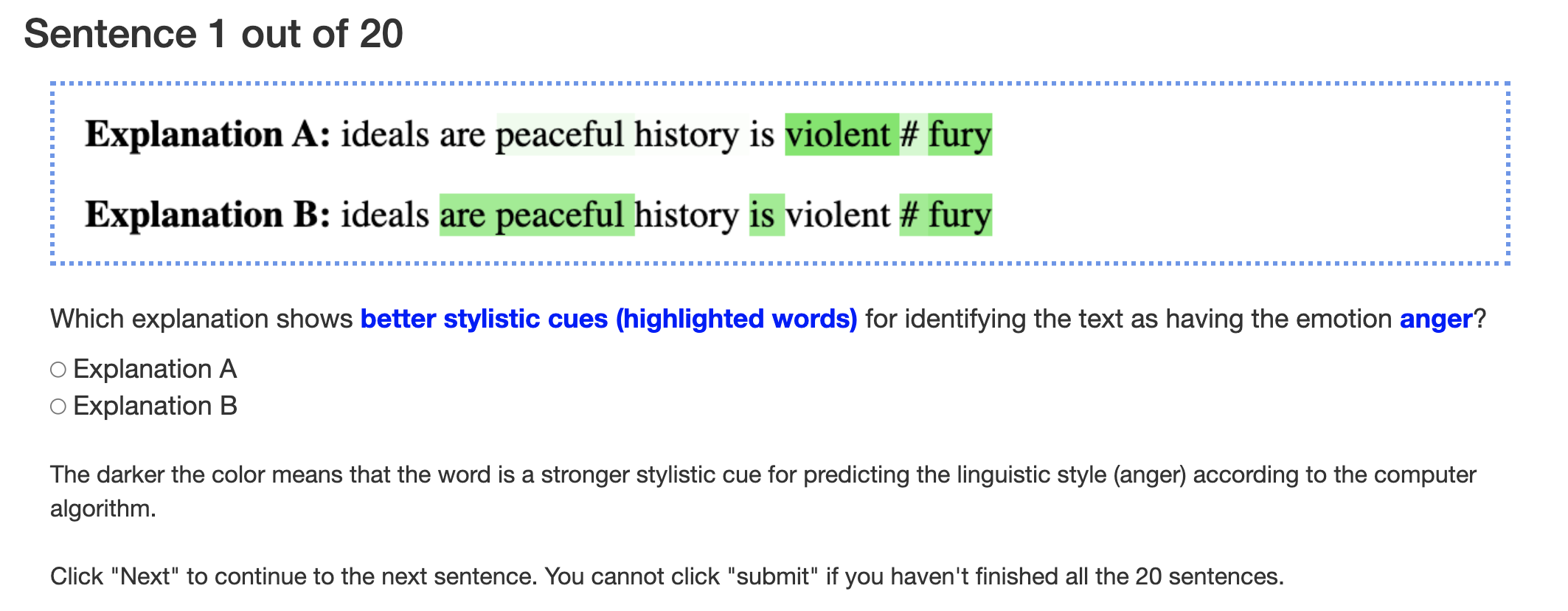}
    \caption{Interface for human evaluation}
    \label{fig:interface}
\end{figure*}

Our interface for human evaluation is shown as in Figure \ref{fig:interface}. Table \ref{tab:ablation_lm} shows results for other pretrained language models. We report word-level style predictor performance tested on \textsc{Hummingbird} test data in the form of Pearson's $r$ correlation scores as follows in Table \ref{tab:extra_ood}. 

\begin{table}[h!]
    \centering
    \setlength{\tabcolsep}{4pt}
    \begin{tabular}{l| c}
       \multirow{2}{*}{\textbf{Style}}
         & \textbf{Word-level}\\
        &  \textbf{Pearson's $r$}\\ \hline
        Politeness &   0.41\\
        Sentiment & 0.61\\
        Offensiveness & 0.37\\
        Anger & 0.45\\
        Disgust & 0.43\\
        Fear & 0.20\\
        Joy  & 0.37\\
        Sadness & 0.41\\
    \end{tabular}
    \caption{Pearson's $r$ scores from training the word-level predicion model and testing it on \textsc{Hummingbird}.}
    \label{tab:extra_ood}
\end{table}

\begin{table*}[]
    \centering
    \setlength{\tabcolsep}{4pt}
    \begin{tabular}{@{}l| c c | c c c | c@{}}
    \toprule
        \textbf{} &  
        \multicolumn{2}{c}{\textbf{\textsc{Hummingbird}}} & \multicolumn{3}{c}{\textbf{\textsc{Original}}} & \textbf{\textsc{OOD}}\\
        Styles$\downarrow$ & Train & Test & Train & Dev & Test & Test\\
        \hline
        Politeness
        
        & 256 (38\%) & 64 (28\%) & 9,855 (55\%) & 530 (56\%) & 567 (57\%) & 1,000 (50\%) \\
        Sentiment 
        & 312 (30\%) & 79 (37\%) & 117,219 (55\%) & 825 (51\%) & 1,749 (50\%) & 5,200 (50\%)\\
        Offensiveness
        & 400 (34\%) & 100 (32\%) & 20,680 (82\%) & 1,173 (82\%) & 1,159 (81\%) & 5,993 (50\%)\\ \hline
        Anger 
        & 400 (35\%) & 100 (34\%) & 6,838 (37\%) & 886 (36\%) & 3,259 (34\%) & 16,168 (50\%) \\
        Disgust & 400 (43\%) & 100 (38\%) & 6,838 (38\%) & 886 (36\%) & 3,259 (34\%) & 10,602 (50\%) \\
        Fear  & 400 (17\%) & 100 (13\%) & 6,838 (18\%) & 886 (14\%) & 3,259 (15\%) & 6,394 (50\%) \\
        Joy & 400 (24\%) & 100 (19\%) & 6,838 (36\%) & 886 (45\%) & 3,259 (44\%) & 15,966 (50\%) \\
        Sadness & 400 (29\%) & 100 (17\%) & 6,838 (29\%) & 886 (30\%) & 3,259 (29\%) & 13,516 (50\%) \\
        % \hline
        % \multicolumn{4}{c}{\textbf{\textsc{Original}}}\\
        % \hline
        % Politeness & 9,855 & 530 & 567 \\
        % Sentiment & 117,219 & 825 & 1749 \\
        % Offensiveness & 20,680 & 1173 & 1159 \\
        % Emotions & 6,838 & 886 & 3259 \\
        % \hline
        % \multicolumn{4}{c}{\textbf{Out-of-Domain}}\\
        % \hline
        % Politeness & \multirow{8}{*}{-} & & 1000 \\
        % Sentiment & & & 5410 \\
        % Offensiveness &  &  & 5993 \\
        % Joy &  &  & 15966 \\ 
        % Sadness & & & 13516 \\
        % Fear & & & 6394 \\
        % Disgust & & & 10602 \\
        % Anger & & & 16168 \\
    \bottomrule
    \end{tabular}
    \caption{Dataset statistics in our experiments. 
    Note that these datasets are preprocessed from existing datasets.
    For \textsc{Hummingbird} \cite{hayati-etal-2021-bert} and \textsc{Original} datasets, the train, dev, and test sets have the same size for all emotions. We do not report the training size of Out-of-Domain (\textsc{OOD}) datasets since we are not using them for training. The label distributions for positive labels are in the parentheses.}
    \label{tab:data_stats}
\end{table*}

\begin{table*}[]
    \centering
    \begin{tabular}{cc|cccccccc}
    \toprule
        \multicolumn{2}{c|}{\multirow{2}{*}{\textbf{Model}}} & \multicolumn{8}{c}{\textbf{F1 Score}} \\
        & & \textbf{Polite.} & \textbf{Sent.} & \textbf{Offens.} & \textbf{Anger} & \textbf{Disgust} & \textbf{Fear} & \textbf{Joy} & \textbf{Sad.} \\
        \midrule
        \multicolumn{10}{c}{\textsc{Orig}} \\\hline\hline
        \multirow{2}{*}{BERT} & Baseline & 67.96 & 96.52 & 97.75 & 89.04 & 86.50 & 95.66 & 88.02 & 88.38  \\
        & \method & 65.84 & 96.59 & 97.81 & 89.01 & 86.90 & 95.63 & 88.14 & 88.41\\ \hline
        \multirow{2}{*}{RoBERTa} & Baseline & 65.83 & 96.94 & 96.40 & 89.56 & 87.17 & 95.68 & 88.32 & 88.52 \\
        & \method & 66.05 & 96.59 & 96.55 & 89.39 & 87.16 & 95.67 & 88.39 & 88.89 \\ \hline
        \multirow{2}{*}{XLNet} & Baseline & 64.09 & 96.57 & 96.86 & 88.46 & 86.07 & 95.55 & 86.95 & 87.32  \\
        & \method & 63.69 & 96.54 & 96.38 & 88.20 & 86.32 & 95.44 & 87.33 & 87.90 \\ \hline
        \multirow{2}{*}{T5} & Baseline & 65.75 & 97.13 & 97.21 & 88.79 & 86.29 & 95.39 & 88.18 & 87.68 \\
        & \method & 67.69 & 97.21 & 96.61 & 88.69 & 86.24 & 95.28 & 87.86 & 87.51 \\
        \midrule
        \multicolumn{10}{c}{\textsc{OOD}} \\\hline\hline
        \multirow{2}{*}{BERT} & Baseline & 71.45 & 86.70 & 88.62 & 77.49 & 74.06 & 78.42 & 75.20 & 78.37 \\
        & \method & 74.18 & 86.99 & 88.98 & 77.51 & 74.63 & 78.48 & 74.26 & 78.71 \\ \hline
        \multirow{2}{*}{RoBERTa} & Baseline & 72.08 & 90.41 & 87.48 & 77.01 & 74.56 & 79.95 & 74.63 & 80.24 \\
        & \method & 69.27 & 89.90 & 89.63 &77.86 &75.07 & 78.66 & 74.18 & 79.09  \\ \hline
        \multirow{2}{*}{XLNet} & Baseline & 68.84 & 88.25 & 88.33 & 76.24 & 74.77 & 78.48 & 74.03 & 78.78\\
        & \method & 67.28 & 89.65 & 88.56 & 76.41 & 74.71 & 78.92 & 74.09 & 78.28  \\ \hline
        \multirow{2}{*}{T5} & Baseline & 70.76 & 91.72 & 88.14 & 75.74 & 73.83 & 79.58 & 73.76 & 78.37 \\
        & \method & 68.14 & 91.73 & 88.05 & 75.48 & 73.33 & 80.23 & 73.70 & 77.61  \\
    \bottomrule
    \end{tabular}
    \caption{More classification results with several language models.}
    \label{tab:ablation_lm}
\end{table*}

\end{document}